\documentclass[11pt,a4paper]{article}
\usepackage{acl2015}
\usepackage{times}
\usepackage{url}
\usepackage{latexsym}
\usepackage{xspace}
\usepackage{amsmath}
\usepackage{amsfonts}
\usepackage{amssymb}
\usepackage{graphicx}
\usepackage{xcolor}

\newcommand{\fb}{{\sf Freebase}\xspace}
\newcommand{\wq}{{\sf WebQuestions}\xspace}
\newcommand{\rv}{{\sf Reverb}\xspace}
\newcommand{\fbq}{{\sf SimpleQuestions}\xspace}
\newcommand{\fbs}{{\sf FB2M}\xspace}
\newcommand{\fbb}{{\sf FB5M}\xspace}
\newcommand{\prp}{{\sf Paraphrases}\xspace}
\newcommand{\wk}{{\sc WikiAnswers}\xspace}

\newcommand{\tofact}[1]{{\tt {\small #1}}}

\renewcommand{\Re}{\mathbb{R}}
\newcommand{\phifact}{f}
\newcommand{\phiq}{g}
\newcommand{\phirv}{h}
\newcommand{\fact}{y}
\newcommand{\rvfact}{y}
\newcommand{\sub}{s}
\newcommand{\rel}{r}
\newcommand{\obj}{o}
\newcommand{\ques}{q}
\newcommand{\score}{S}

\title{Large-scale Simple Question Answering with Memory Networks}

\author{
Antoine Bordes\\
  Facebook AI Research \\
770 Broadway\\
New York, NY. USA\\
  {\tt abordes@fb.com} \\
\And
Nicolas Usunier\\
Facebook AI Research \\
 112, avenue de Wagram\\
75017 Paris, France\\
  {\tt usunier@fb.com} \\
\And
Sumit Chopra, Jason Weston\\
  Facebook AI Research \\
770 Broadway\\
New York, NY. USA\\
  {\tt \{spchopra, jase\}@fb.com} \\
}

\date{}

\begin{document}
\maketitle
\begin{abstract}
  Training large-scale question answering systems is complicated
  because training sources usually cover a small portion of the range
  of possible questions.
  This paper studies the impact of multitask and transfer learning for
  {\it simple question answering}; a setting for which the reasoning
  required to answer is quite easy, as long as one can retrieve the
  correct evidence given a question, which can be difficult in
  large-scale conditions.
  To this end, we introduce a new dataset of 100k questions that we
  use in conjunction with existing benchmarks.
  We conduct our study within the framework of Memory Networks
  \cite{weston2014memory} because this perspective allows us to eventually
  scale up to more complex reasoning, and show that Memory Networks
  can be successfully trained to achieve excellent performance.
\end{abstract}

\begin{table*}[t]
\begin{center}
\resizebox{1\linewidth}{!}{
  \begin{tabular}{@{}l@{}r@{}}
    What American cartoonist is the creator of Andy Lippincott? &
                                                                   \tofact{(andy\_lippincott,
                                                                  character\_created\_by,
                                                                   \underline{garry\_trudeau})} \\
    Which forest is Fires Creek in? & \tofact{(fires\_creek,
                                       containedby,
                                       \underline{nantahala\_national\_forest})} \\
    What is an active ingredient in childrens earache relief ? &
                                                                 \tofact{(childrens\_earache\_relief,
                                                                 active\_ingredients,
                                                                 \underline{capsicum)}} \\
    What does Jimmy Neutron do? & \tofact{(jimmy\_neutron,
                                 fictional\_character\_occupation, \underline{inventor})}\\
    What dietary restriction is incompatible with kimchi? &
                                                           \tofact{(kimchi, incompatible\_with\_dietary\_restrictions, \underline{veganism})}
  \end{tabular}
}
\caption{\label{tab:ex} {\bf Examples of simple QA}. Questions and
  corresponding facts have been extracted from the new dataset \fbq
  introduced in this paper. Actual answers are underlined.}
\end{center}
\vspace*{-2ex}
\end{table*}

\section{Introduction}

Open-domain Question Answering (QA) systems aim at providing the exact
answer(s) to questions formulated in natural language, without
restriction of domain. While there is a long history of QA systems
that search for textual documents or on the Web and extract answers
from them (see e.g. \cite{voorhees2000overview,dumais2002web}), recent
progress has been made with the release of large Knowledge Bases (KBs)
such as \fb, which contain consolidated knowledge stored as atomic
facts, and extracted from different sources, such as free text, tables
in webpages or collaborative input.
%
%
Existing approaches for QA from KBs use learnable components to either
transform the question into a structured KB query
\cite{berant-EtAl:2013:EMNLP} or learn to embed questions and facts in
a low dimensional vector space and retrieve the answer by computing
similarities in this embedding space
\cite{bordes-chopra-weston:2014:EMNLP2014}. However, while most recent
efforts have focused on designing systems with higher reasoning
capabilities, that could jointly retrieve and use multiple facts to
answer, the simpler problem of answering questions that refer to a
single fact of the KB, which we call {\it Simple Question Answering}
in this paper, is still far from solved.

Hence, existing benchmarks are small; they mostly cover the
head of the distributions of facts, and are restricted in their
question types and their syntactic and lexical variations. As such, it
is still unknown how much the existing systems perform outside the
range of the specific question templates of a few, small benchmark
datasets, and it is also unknown whether learning on a single dataset
transfers well on other ones, and whether such systems can learn from
different training sources, which we believe is necessary to capture
the whole range of possible questions.

Besides, the actual need for reasoning, i.e. constructing the answer
from more than a single fact from the KB, depends on the actual
structure of the KB. As we shall see, for instance, a simple
preprocessing of \fb tremendously increases the coverage of simple QA
in terms of possible questions that can be answered with a single
fact, including list questions that expect more than a single answer. In
fact, the task of simple QA itself might already cover a wide range of
practical usages, if the KB is properly organized.

This paper presents two contributions. First, as an effort to study
the coverage of existing systems and the possibility to train jointly
on different data sources via multitasking, we collected the first
large-scale dataset of questions and answers based on a KB, called
\fbq. This dataset, which is presented in Section \ref{sec:fbq},
contains more than $100$k questions written by human annotators and
associated to \fb facts, while the largest existing benchmark, \wq,
contains less than $6$k questions created automatically using the
Google suggest API.

Second, in sections \ref{sec:memnn} and \ref{sec:training}, we present
an embedding-based QA system developed under the framework of Memory
Networks (MemNNs) \cite{weston2014memory,sukhbaatar2015weakly}. Memory
Networks are learning systems centered around a memory component that
can be read and written to, with a particular focus on cases where the
relationship between the input and response languages (here natural
language) and the storage language (here, the facts from KBs) is
performed by embedding all of them in the same vector space. The
setting of the simple QA corresponds to the elementary operation of
performing a single lookup in the memory. While our model bares
similarity with previous embedding models for QA
\cite{bordes2014open,bordes-chopra-weston:2014:EMNLP2014}, using the
framework of MemNNs opens the perspective to more involved inference
schemes in future work, since MemNNs were shown to perform well on
complex reasoning toy QA tasks \cite{weston2014memory}. We discuss
related work in Section \ref{sec:related}.

We report experimental results in Section \ref{sec:expes},
where we show that our model achieves excellent results on the
benchmark \wq. We also show that it can learn from two different QA
datasets to improve its performance on both. We also present the first
successful application of transfer learning for QA. Using the \rv KB
and QA datasets, we show that \rv facts can be added to the memory and
used to answer {\it without retraining}, and that MemNNs achieve
better results than some systems designed on this dataset.

\section{Simple Question Answering}
\label{sec:fbq}

Knowledge Bases contain facts expressed as triples \tofact{(subject,
  relationship, object)}, where \tofact{subject} and \tofact{object}
are entities and \tofact{relationship} describes the type of
(directed) link between these entities. 
The simple QA problem we address here consist in finding the
answer to questions that can be rephrased as queries of the form
\tofact{(subject, relationship, ?)}, asking for all objects linked to
\tofact{subject} by \tofact{relationship}. The question {\it What do
  Jamaican people speak ?}, for instance, could be rephrased as the
\fb query \tofact{(jamaica, language\_spoken, ?)}.
In other words, fetching a single fact from a KB is sufficient to
answer correctly.

The term {\it simple QA} refers to the simplicity of the reasoning
process needed to answer questions, since it involves a single fact.
However, this does not mean that the QA problem is easy per se, since
retrieving this single supporting fact can be very challenging as
it involves to search over millions of alternatives given a query
expressed in natural language.
Table~\ref{tab:ex} shows that, with a KB with many
types of relationships like \fb, the range of questions that can be
answered with a single fact is already very broad.
Besides, as we shall see, modiying slightly the structure
of the KB can {\it make some QA problems simpler} by adding direct
connections between entities and hence allow to bypass the need for
more complex reasoning.

\subsection{Knowledge Bases}


We use the KB \fb\footnote{\url{www.freebase.com}} as
the basis of our QA system, our source of facts and answers.  
All \fb entities and relationships are typed and the lexicon for types and
relationships is closed. \fb data is collaboratively collected and
curated, to ensure a high reliability of the facts. Each entity
has an internal identifier and a set of strings that are usually used
to refer to that entity in text, termed {\it aliases}.
We consider two extracts of \fb,  whose statistics are given in
Table~\ref{tab:kbs}. \fbs, which was used
in \cite{bordes-chopra-weston:2014:EMNLP2014}, contains about $2$M
entities and $5$k relationships. \fbb, is much larger with about $5$M entities
and more than $7.5$k relationships.

We also use the KB \rv 
as a
secondary source of facts to study how well a model trained to answer
questions using \fb facts could be used to answer using \rv's as well,
without being trained on \rv data. This is a pure setting of {\it transfer
  learning}.
\rv is interesting for this experiment because it differs a lot from \fb.
Its data was extracted automatically from text with minimal human
intervention and is highly unstructured: entities are unique strings and the lexicon
for relationships is open. This leads to many more relationships, but
entities with multiple references are not deduplicated, ambiguous
referents are not resolved, and the reliability of the stored facts is
much lower than in \fb.
We used the full extraction from \cite{ReVerb2011}, which contains
$2$M entities and $600$k relationships.

\begin{table}
\begin{center}
\resizebox{1\linewidth}{!}{
\begin{tabular}{|l|r|r|r|}
\cline{2-4}
\multicolumn{1}{c|}{} & \multicolumn{1}{c|}{\fbs} & \multicolumn{1}{c|}{\fbb} & \multicolumn{1}{c|}{\rv} \\
\hline
{\sc Entities} & 2,150,604 & 4,904,397 & 2,044,752 \\
{\sc Relationships} &  6,701 & 7,523 & 601,360 \\
{\sc Atomic facts} & 14,180,937 & 22,441,880 & 14,338,214 \\
{\sc Facts} (grouped) &  10,843,106 & 12,010,500 & --\\
\hline
\end{tabular}
}
\caption{\label{tab:kbs} {\bf Knowledge Bases} used in this
  paper. \fbs and \fbb are two versions of \fb.}
\end{center}
\vspace*{-1ex}
\end{table}

\subsection{The SimpleQuestions dataset}

Existing resources for QA such as \wq \cite{berant-EtAl:2013:EMNLP} are rather
small (few thousands questions) and hence do not provide a very
thorough coverage of  the variety of questions that could be answered
using a KB like \fb, even in the context of simple QA.
Hence, in this paper, we introduce a new dataset of much larger scale
for the task of simple QA called \fbq.\footnote{The dataset is
  available from \url{http://fb.ai/babi}.}
This dataset consists of a total of 108,442 questions written in
natural language by human English-speaking annotators each paired with
a corresponding fact from \fbs that provides the answer and explains it.
We randomly shuffle these questions and use 70\% of them (75910) as
training set, 10\% as validation set (10845), and the remaining 20\%
as test set.
Examples of questions and facts are given in Table~\ref{tab:ex}.

\iftrue
We collected \fbq in  two phases.
The first phase consisted of shortlisting the set of facts from \fb
to be annotated with questions. 
We used \fbs as background KB and removed
all facts with undefined relationship type i.e. containing the word
\tofact{freebase}. We also removed all facts for which the
(subject, relationship) pair had more than a threshold number
of objects. This filtering step is crucial to remove
facts which would result in trivial uninformative questions, such as,
{\it Name a person who is an actor?}. The threshold was set to 10. 

In the second phase, these selected facts were sampled and delivered
to human annotators to generate questions from them.
For the sampling, each fact was associated with a probability which
defined as a function of its relationship frequency in the KB: to
favor variability, facts with relationship appearing more frequently were
given lower probabilities.
For each sampled facts, annotators were shown the facts along with
hyperlinks to \url{freebase.com} to provide some
context while framing the question. Given this information, annotators were asked to
phrase a question involving the subject and the
relationship of the fact, with the answer being the object.
The annotators were explicitly instructed to phrase the question differently as much as
possible, if they encounter multiple facts with similar relationship.
They were also given the option of skipping facts if they wish to do so.
This was very important to avoid the annotators
to write a boiler plate questions when they had no background
knowledge about some facts.
\fi

\section{Memory Networks for Simple QA}
\label{sec:memnn}
A Memory Network consists of a memory (an indexed array of objects)
and a neural network that is trained to query it given some inputs
(usually questions). It has four components: {\it Input map} ($I$), {\it
  Generalization} ($G$), {\it Output map} ($O$) and {\it Response} ($R$)
which we detail below.
%
But first, we describe the MemNNs workflow used to set up a model for
simple QA. This proceeds in three steps:
\paragraph{1. Storing Freebase:} this first phase parses \fb (either \fbs
  or \fbb depending on the setting) and stores it in memory. It uses
  the {\it Input} module to preprocess the data.
\paragraph{2. Training:} this second phase trains the MemNN to answer
  question. This uses {\it Input}, {\it Output} and {\it Response}
  modules, the training concerns mainly the parameters of the
  embedding model at the core of the  {\it Output} module.
\paragraph{3. Connecting Reverb:}this third phase adds new facts coming
  from \rv to the memory. This is done after training to test the
  ability of MemNNs to handle new facts without having to be
  re-trained. It uses the {\it Input} module to preprocess \rv facts and the {\it
    Generalization} module to connect them to the facts already
  stored.

After these three stages, the MemNN is ready to answer any question by
running the $I$, $O$ and $R$ modules in turn. 
We now detail the 
implementation of the four modules.

\subsection{Input module}
\label{sec:MemNNinput}
This module preprocesses the three types of data that are
input to the network: \fb facts that are used to populate the memory,
questions that the system need to answer, and \rv facts
that we use, in a second phase, to extend the memory.

\paragraph{Preprocessing Freebase}
The \fb data is initially stored as atomic facts involving single
entities as subject and object, plus a relationship between
them. However, this storage needs to be adapted to the QA task in two
aspects.

First, in order to answer list questions, which expect more than one
answer, we redefine a fact as being a triple containing a subject, a
relationship, and the set of all objects linked to the subject by the
relationship. This {\it grouping} process transforms atomic facts
into grouped facts, which we simply refer to as {\it facts} in the
following.
Table~\ref{tab:kbs} shows the impact of this grouping: on \fbs, this
decreases the number of facts from $14$M to $11$M and, on \fbb, from
$22$M to $12$M.

Second, the underlying structure of \fb is a hypergraph, in which
more than two entities can be linked. For instance dates can be linked
together with two entities to specify the time period over which the
link was valid. The underlying triple storage involves {\it mediator
  nodes} for each such fact, effectively making entities linked
through paths of length 2, instead of 1. To obtain direct
links between entities in such cases, we created a single fact for
these facts by removing the intermediate node and using the
second relationship as the relationship for the new condensed fact. This
step reduces the need for searching the answer outside the immediate
neighborhood of the subject referred to in the question, widely
increasing the scope of the simple QA task on \fb.
On \wq, a benchmark not primarily designed for simple
QA, removing mediator nodes allows to jump from around $65$\% to
$86$\% of questions that can be answered with a single fact.

\paragraph{Preprocessing Freebase facts}
A fact with $k$ objects $\fact = (\sub, \rel, \{\obj_1, ...,
\obj_k\})$ is represented by a bag-of-symbol vector $\phifact(\fact)$
in $\Re^{N_S}$, where $N_S$ is the number of entities and
relationships. Each dimension of $\phifact(\fact)$ corresponds
to a relationship or an entity (independent of whether it appears as
subject or object). The entries of the subject and of the
relationship have value $1$, and the entries of the objects are set to
$1/k$. All other entries are $0$.

\paragraph{Preprocessing questions}
A question $\ques$ is mapped to a bag-of-ngrams representation
$\phiq(\ques)$ of dimension $\Re^{N_V}$ where $N_V$ is the size of the
vocabulary. The vocabulary contains all individual words that appear
in the questions of our datasets, together with the aliases of \fb
entities, each alias being a single n-gram. The entries of
$\phiq(\ques)$ that correspond to words and n-grams of $\ques$ are
equal to $1$, all other ones are set to $0$.


\paragraph{Preprocessing Reverb facts}
In our experiments with \rv, each fact $\rvfact = (\sub, \rel, \obj)$
is represented as a vector $h(\rvfact)\in\Re^{N_S+N_V}$. This vector is a
bag-of-symbol for the subject $\sub$ and the object $\obj$, and a
bag-of-words for the relationship $\rel$. The exact composition of $h$
is
provided by the {\it Generalization} module, which we describe now.

\subsection{Generalization module}

This module is responsible for adding new elements to the memory. In
our case, the memory has a multigraph structure where each node is a
\fb entity and labeled arcs in the multigraph are \fb relationships:
after their preprocessing, all \fb facts are stored using this structure.

We also consider the case where new facts, with a different structure
(i.e. new kinds of relationship), are provided to the MemNNs by using
\rv.
In this case, the generalization module is then used to connect \rv
facts to the \fb-based memory structure, in order to make them usable
and searchable by the MemNN.

To link the subject and the object of a \rv fact to \fb entities, we
use precomputed entity
links 
\cite{lin2012entity}. If such links do not give any result for an
entity, we search for \fb entities with at least one alias that
matches the \rv entity string. These two processes allowed to match
$17$\% of \rv entities to \fb ones. The remainder of entities were
encoded using bag-of-words representation of their strings, since we
had no other way of matching them to \fb entities.
All \rv relationships were encoded using bag-of-words of
their strings.
Using this approximate process, we are able to store each \rv fact as
a bag-of-symbols (words or \fb entities) all already seen by the MemNN
during its training phase based on \fb.
We can then hope that what had been learned there could also be
successfully used to query \rv facts.

\subsection{Output module}
\label{sec:outputModule}
The output module performs the memory lookups given the input to
return the {\it supporting facts} destined to eventually provide the
answer given a question.
In our case of simple QA, this module only returns a single supporting fact.
To avoid scoring all the stored
facts, we first perform an approximate entity linking step to generate
a small set of candidate facts. The supporting fact is the candidate fact that
is most similar to the question according to an embedding model.

\paragraph{Candidate generation}
To generate candidate facts, we match $n$-grams of words of the question to
aliases of \fb entities and select a few matching entities. All facts
having one of these entities as subject are scored in a second step.

We first generate all possible $n$-grams from the question, removing
those that contain an interrogative pronoun or $1$-grams
that belong to a list of stopwords. We only keep the $n$-grams which
are an alias of an entity, and then discard all $n$-grams that are a
subsequence of another $n$-gram, except if the longer $n$-gram only
differs by {\it in}, {\it of}, {\it for} or {\it the} at the
beginning. We finally keep the two entities with the most links in \fb
retrieved for each of the five longest matched $n$-grams.

\paragraph{Scoring}
Scoring is performed using an embedding model. Given two embedding
matrices ${\bf W}_V \in \Re^{d\times N_V}$ and ${\bf W}_S \in
\Re^{d\times N_S}$, which respectively contain, in columns, the
$d$-dimensional embeddings of the words/$n$-grams of the vocabulary
and the embeddings of the \fb entities and relationships, the
similarity between question $\ques$ and a \fb candidate fact $\fact$ is
computed as:
\begin{equation*}
\score_{QA}(\ques, \fact) = \cos({\bf W}_V \phiq(\ques), {\bf
  W}_S\phifact(\fact))\,,
\end{equation*}
with $ \cos()$ the cosine similarity.
When scoring a fact $\rvfact$ from \rv, we use the same embeddings and
build the matrix ${\bf W}_{VS} \in \Re^{d\times (N_V+N_S)}$, which
contains the concatenation in columns of ${\bf W}_V$ and ${\bf W}_S$,
and also compute the cosine similarity:
\begin{equation*}
\score_{RVB}(\ques, \rvfact) = \cos({\bf W}_V \phiq(\ques), {\bf W}_{VS}\phirv(\rvfact))\,.
\end{equation*}
The dimension $d$ is a hyperparameter, and the embedding
matrices ${\bf W}_V$ and ${\bf W}_S$ are the parameters learned
with the training algorithm of Section~\ref{sec:training}.

\subsection{Response module}
In Memory Networks, the {\it Response} module post-processes the
result of the {\it Output} module to compute the intended answer. In
our case, it returns the set of objects of the selected supporting
fact.

\section{Training}
\label{sec:training}

This section details how we trained the scoring function of the {\it
  Output} module using a multitask training process on
four different sources of data. 

First, in addition to the new \fbq dataset described in Section
\ref{sec:fbq}, we also used \wq, a benchmark for QA 
introduced in \cite{berant-EtAl:2013:EMNLP}: questions are labeled with answer
strings from aliases of \fb entities, and many questions expect
multiple answers.
Table~\ref{tab:data} details the statistics of both datasets.

We also train on automatic questions generated from the KB, that is
\fbs or \fbb depending on the setting, which are essential to learn
embeddings for the entities not appearing in either \wq or \fbq.
Statistics of \fbs or \fbb are given in Table~\ref{tab:kbs}; we
generated one training question per fact following the same process 
as that used in \cite{bordes-chopra-weston:2014:EMNLP2014}.

Following previous work such as \cite{paralex}, we also use
the indirect supervision signal of pairs of question paraphrases.
We used a subset of the large set of paraphrases extracted
from \wk and introduced in \cite{fader2014open}.
Our \prp dataset is made of $15$M clusters containing 2 or more
paraphrases each.

\subsection{Multitask training}

As in previous work on embedding models and Memory Networks
\cite{bordes-chopra-weston:2014:EMNLP2014,bordes2014open,weston2014memory},
the embeddings are trained with a ranking criterion. For QA datasets the
goal is that in the embedding space, a supporting fact is more similar to
the question than any other {\it non-supporting} fact. For the paraphrase dataset, a
question should be more similar to one of its paraphrases than to any
another question.

The multitask learning of the embedding matrices ${\bf W}_V$ and ${\bf
  W}_S$ is performed by alternating stochastic gradient descent (SGD)
steps over the loss function on the different datasets. For the QA
datasets, given a question/supporting fact pair $(\ques, \fact)$ and a
non-supporting fact $\fact'$, we perform a step to minimize the loss function
\begin{equation*}
\ell_{QA}(\ques, \fact, \fact') = \big[\gamma - \score_{QA}(\ques, \fact) + \score_{QA}(\ques, \fact') \big]_+\,,
\end{equation*}
where $[.]_+$ is the positive part and $\gamma$ is a margin
hyperparameter. For the paraphrase dataset, the similarity score
between two questions $\ques$ and $\ques'$ is also the cosine between
their embeddings, i.e. $\score_{QQ}(\ques, \ques') = \cos({\bf W}_V
\phiq(\ques), {\bf W}_{V}\phiq(\ques'))$, and given a paraphrase pair
$(\ques, \ques')$ and another question $\ques''$, the loss is:
\begin{equation*}
\ell_{QQ}(\ques, \ques', \ques'') = \big[\gamma - \score_{QQ}(\ques, \ques') + \score_{QQ}(\ques, \ques'') \big]_+\,.
\end{equation*}
The embeddings (i.e. the columns of ${\bf W}_V$ and ${\bf W}_S$) are
projected onto the $L_2$ unit ball after each update.
At each time step, a sample from the paraphrase dataset is drawn with
probability $0.2$ (this probability is arbitrary). Otherwise, a sample from one of the three
QA datasets, chosen uniformly at random, is taken. We use the WARP loss
\cite{wsabie} to speed up training, and Adagrad
\cite{duchi2011adaptive} as SGD algorithm multi-threaded with {\tt
  HogWild!} \cite{recht2011hogwild}.
Training takes 2-3 hours on 20 threads.

\subsection{Distant supervision}
Unlike for \fbq or the synthetic QA data generated from \fb,  for
\wq only answer strings are provided for questions: the
supporting facts are unknown.

In order to generate the supervision, we use the candidate fact
generation algorithm of Section~\ref{sec:outputModule}. For each
candidate fact, the aliases of its objects are compared to the set of
provided answer strings. The fact(s) which can generate the maximum
number of answer strings from their objects' aliases are then kept. If
multiple facts are obtained for the same question, the ones with
the minimal number of objects are considered as supervision facts. This
last selection avoids favoring  irrelevant relationships that would be
kept only because they point to many objects but would not be specific
enough. If no answer string could be found from the objects of the
initial candidates, the question is discarded from the training set.

Future work should investigate the process of weak supervised training
of MemNNs recently introduced in \cite{sukhbaatar2015weakly} that
allows to train them without any supervision coming from the
supporting facts.

\subsection{Generating negative examples}
As in \cite{bordes-chopra-weston:2014:EMNLP2014,bordes2014open},
learning is performed with gradient descent, so that negative examples
(non-supporting facts or non-paraphrases) are generated according to a
randomized policy during training.

For paraphrases, given a pair $(q, q')$, a non-paraphrase
pair is generated as $(q, q'')$ where $q''$ is a random question of
the dataset, not belonging to the cluser of $q$.
For question/supporting fact pairs, we use two policies. The
default policy to obtain a non-supporting fact is to corrupt the
answer fact by exchanging its subject, its relationship or its
object(s) with that of another fact chosen uniformly at random from
the KB. In this policy, the element of the fact to corrupt is chosen
randomly, with a small probability (0.3) of corrupting more than one element
of the answer fact. The second policy we propose, called {\it
  candidates as negatives}, is to take as non-supporting fact a
randomly chosen fact from the set of candidate facts. While the first
policy is standard in learning embeddings, the second one is more
original, and, as we see in the experiments, gives slightly
better performance.

\begin{table}
\begin{center}
\resizebox{1\linewidth}{!}{
\begin{tabular}{|l|r|r|r|}
\cline{2-4}
\multicolumn{1}{c|}{} & \multicolumn{1}{c|}{\wq} & \multicolumn{1}{c|}{\fbq} & \multicolumn{1}{c|}{\rv} \\
\hline
{\sc Train} & 3,000 & 75,910 & -- \\
{\sc Valid.} &  778 & 10,845 & -- \\
{\sc Test} & 2,032 & 21,687 & 691 \\
\hline
\end{tabular}
}
  \caption{\label{tab:data} {\bf Training and evaluation datasets}. Questions
    automatically generated from the KB and paraphrases can also be
    used in training.}
\end{center}
\end{table}

\section{Related Work}
\label{sec:related}

The first approaches to open-domain QA were search engine-based
systems, where keywords extracted from the question are sent to a
search engine, and the answer is extracted from the top results
\cite{yahya2012natural,unger2012template}. This method has been
adapted to KB-based QA \cite{yahya2012natural,unger2012template}, and
obtained competitive results with respect to semantic parsing and
embedding-based approaches.

Semantic parsing approaches
\cite{cai-yates:2013:ACL2013,berant-EtAl:2013:EMNLP,kwiatkowski-EtAl:2013:EMNLP,berant2014semantic,fader2014open}
perform a functional parse of the sentence that can be interpreted as
a KB query. Even though these approaches are difficult to train at
scale because of the complexity of their inference, their advantage is
to provide a deep interpretation of the question. Some of these
approaches require little to no question-answer pairs
\cite{paralex,reddy2014large}, relying on simple rules to tranform the
semantic interpretation to a KB query.

Like our work, embedding-based methods for QA can be seen as simple
MemNNs. The algorithms of \cite{bordes2014open,weston2014memory} use
an approach similar to ours but are based on \rv rather than \fb, and
relied purely on bag-of-word for both questions and facts. The
approach of \cite{yang2014joint} uses a different representation of
questions, in which recognized entities are replaced by an {\it
  entity} token, and a different training data using entity
mentions from {\sc Wikipedia}. Our model is closest to the one presented
in \cite{bordes-chopra-weston:2014:EMNLP2014}, which is discussed in
more details in the experiments.

\section{Experiments}
\label{sec:expes}
This section provides an extensive evaluation of our MemNNs
implementation against state-of-the-art QA methods as well as an
empirical study of the impact of using multiple training sources on
the prediction performance.

\subsection{Evaluation and baselines}
 Table~\ref{tab:data} details the dimensions of the test sets of \wq, \fbq
 and \rv which we used for evaluation.
 On \wq, we evaluate against previous results on this benchmark
 \cite{berant-EtAl:2013:EMNLP,yao2014information,berant2014semantic,bordes-chopra-weston:2014:EMNLP2014,yang2014joint}
 in terms of F1-score as defined in \cite{berant2014semantic}, which is the
 average, over all test questions, of the F1-score of the sets of
 predicted answers.
 Since no previous result was published on \fbq, we only compare
 different versions of MemNNs.  \fbq questions are labeled with their
 entire \fb fact, so we evaluate in terms of path-level
 accuracy, in which a prediction is correct if the subject and the
 relationship were correctly retrieved by the system.

The \rv test set, based on the KB of the same name and introduced in
\cite{paralex} is used for evaluation only. It contains $691$
questions. We consider the task of re-ranking a small set of candidate
answers, which are \rv facts and are labeled as correct or
incorrect. We compare our approach to the original system
\cite{paralex}, to \cite{bordes2014open} and to the original MemNNs
\cite{weston2014memory}, in terms of accuracy, which is the percentage
of questions for which the top-ranked candidate fact is correct.

\subsection{Experimental setup}
All models were trained with at least the dataset made of
synthetic questions created from the KB.
The hyperparameters were chosen to maximize the F1-score on \wq
validation set, independently of the testing dataset. The
embedding dimension and the learning rate were chosen among
$\{64, 128, 256\}$ and $\{1, 0.1, ..., 1.0e-4\}$ respectively, and the
margin $\gamma$ was set to $0.1$.
For each configuration of
hyperparameters, the F1-score on the validation set was computed
regularly during learning to perform early stopping.

We tested additional configurations for our algorithm. First, in the
{\it Candidates as Negatives} setting (negative facts are sampled from
the candidate set, see Section \ref{sec:training}), abbreviated {\sc
  Cands As Negs}, the experimental protocol is the same as in the
default setting but the embeddings are initialized with the best
configuration of the default setup. Second, our model shares some
similarities with an approach studied in
\cite{bordes-chopra-weston:2014:EMNLP2014}, in which the authors
noticed important gains using a subgraph representation of
answers. For completeness, we also added such a subgraph
representation of objects. In that setting, called {\it Subgraph},
each object $\obj$ of a fact is itself represented as a
bag-of-entities that encodes the immediate neighborhood of
$o$. This {\it Subgraph} model is trained similarly as our main
approach and only the results of a post-hoc ensemble combination of the two
models (where the scores are added) are presented.
We also report the results obtained by an ensemble of the 5 best models
on validation (subgraph excepted); this is denoted {\it 5 models}.

\begin{table*}
\begin{center}
\begin{small}
\begin{tabular}{|c|c|c|c|c|c|c|c|c|}
\cline{7-9}
\multicolumn{6}{c|}{} & \wq & \fbq & \rv \\
\multicolumn{6}{c|}{} & {\sc F1-score (\%)} & {\sc Accuracy (\%)} &
                                                                    {\sc
                                                                    Accuracy
                                                                    (\%)}
  \\
\hline
\multicolumn{9}{|l|}{{\sc Baselines}} \\
\hline
\multicolumn{6}{|c|}{ Random guess} & 1.9 & 4.9 & 35 \\
\multicolumn{6}{|c|}{\cite{berant-EtAl:2013:EMNLP}} & 31.3 & n/a & n/a \\
\multicolumn{6}{|c|}{\cite{fader2014open}} & n/a  & n/a & 54 \\
\multicolumn{6}{|c|}{\cite{bordes2014open}} & 29.7  & n/a & {\bf 73}
  \\
\multicolumn{6}{|c|}{\cite{bordes-chopra-weston:2014:EMNLP2014} -- {\it
  using path}} &
                                                                   35.3
                                   & n/a &  n/a \\
\multicolumn{6}{|c|}{\cite{bordes-chopra-weston:2014:EMNLP2014} -- {\it
  using path + subgraph}} &
                                                                   39.2
                                   & n/a &  n/a \\
\multicolumn{6}{|c|}{\cite{berant2014semantic}} & 39.9 & n/a & n/a \\
\multicolumn{6}{|c|}{\cite{yang2014joint}} & 41.3 & n/a & n/a \\
\multicolumn{6}{|c|}{\cite{weston2014memory}  -- {\it the original MemNN}} & n/a & n/a & 72 \\

\hline
\multicolumn{9}{|l|}{{\sc Memory Networks} {\it (never trained on \rv~--
  only transfer)}} \\
\hline
{\sc KB} & \multicolumn{3}{|c|}{\sc Train sources} &
                                           {\sc Cands} & {\sc Ensemble} &
                                                                    \multicolumn{3}{|c|}{}\\
 & {\sf WQ} & {\sf SIQ} & {\sf PRP} &  {\sc As Negs} & & \multicolumn{3}{|c|}{}\\

\hline
\fbs & yes & yes & yes & -- & -- & 36.2 & 62.7 & n/a \\
\hline
\fbb & -- & -- & -- & -- & -- & 18.7 & 44.5 & 52 \\
\fbb & -- & -- & yes & -- & -- & 22.0 & 48.1 & 62 \\
\fbb & -- & yes & -- & -- & -- & 22.7 & 61.6 & 52 \\
\fbb & -- & yes & yes & -- & -- & 28.2 & 61.2 & 64 \\
\fbb & yes & -- & -- & -- & -- & 40.1 & 46.6 & 58 \\
\fbb & yes & -- & yes & -- & -- & 40.4 & 47.4 & 61 \\
\fbb & yes & yes & -- & -- & -- & 41.0 & 61.7 & 52 \\
\fbb & yes & yes & yes & -- & -- & 41.0 & 62.1 & 67 \\
\hline
\fbb & yes & yes & yes & yes & -- & 41.2 & 62.2 & 65 \\
\fbb & yes & yes & yes & yes & 5 models & 41.9 & {\bf 63.9}  & {\it 68} \\
\fbb & yes & yes & yes & yes & Subgraph & {\bf 42.2} & 62.9 & 62 \\
\hline
\end{tabular}
\end{small}
\caption{\label{tab:res} {\bf Experimental results} for previous
  models of the literature and variants of Memory Networks. All
  results are on the test sets. {\sf WQ}, {\sf
    SIQ} and {\sf PRP} stand for \wq, \fbq and \prp respectively. More
details in the text.}
\end{center}
\vspace*{-1ex}
\end{table*}

\subsection{Results}

\paragraph{Comparative results}
The results of the comparative experiments are given in Table
\ref{tab:res}. 
On the main benchmark \wq, our best results use all data sources, the
bigger extract from \fb and the {\sc Cands As Negs} setting. The two
ensembles achieve excellent results, with F1-scores of $41.9\%$
and $42.2\%$ respectively. The best published competing approach
\cite{yang2014joint} has an F1-score of $41.3\%$, which is comparable
to a single run of our model ($41.2\%$). On the new \fbq dataset, the
best models achieve $62-63\%$ accuracy, while the supporting fact is
in the candidate set for about $86\%$ of \fbq questions. This shows
that MemNNs are effective at re-ranking the candidates, but also that
simple QA is still not solved.


Our approach bares similarity to
\cite{bordes-chopra-weston:2014:EMNLP2014} - {\it using path}. They
use \fbs, and so their result ($35.3\%$ F1-score on \wq) should be
compared to our $36.2\%$. The models are slightly different in that
they replace the entity string with the subject entity in the question
representation and that we use the cosine similarity instead of the
dot product, which gave consistent improvements. Still, the major
differences come from how we use \fb. First, the removal of the
mediator nodes allows us to restrict ourselves to single supporting
facts, while they search in paths of length 2 with a heuristic to
select the paths to follow (otherwise, inference is too costly), which
makes our inference simpler and more efficient. Second, using grouped
facts, we integrate multiple answers during learning (through the
distant supervision), while they use a grouping heuristic at test
time. Grouping facts also allows us to scale much better and to train
on \fbb. On \wq, not specifically designed as a simple QA dataset,
$86\%$ of the questions can now be answered with a single supporting
fact, and performance increases significantly (from $36.2\%$ to
$41.0\%$ F1-score). Using the bigger \fbb as KB does not change performance on \fbq
because it was based on \fbs, but the results show that our model is
robust to the addition of more entities than necessary.



\paragraph{Transfer learning on Reverb}

In this set of experiments, all \rv facts are added to the memory,
without any retraining, and we test our ability to rerank answers on
the companion QA set. Thus, Table \ref{tab:res} (last column) presents
the result of our model {\it without training} on \rv~ {\it against
  methods specifically developed on that dataset}. Our best results
are $67\%$ accuracy (and $68\%$ for the ensemble of $5$ models), which
are better than the $54\%$ of the original paper and close to the
state-of-the-art $73\%$ of \cite{bordes2014open}. These results show
that the Memory Network approach can integrate and use new entities and
links.



\paragraph{Importance of data sources}

The bottom half of Table \ref{tab:res} presents the results on the
three datasets when our model is trained with different data
sources. We first notice that models trained on a single QA dataset
perform poorly on the other datasets (e.g. $46.6\%$ accuracy on \fbq for the
model trained on \wq only), which shows that the performance on \wq
does not necessarily guarantee high coverage for simple QA. On the
other hand, training on both datasets only improves performance; in
particular, the model is able to capture all question patterns of the
two datasets; there is no ``negative interaction''.

While paraphrases do not seem to help much on \wq and \fbq, except
when training only with synthetic questions, they have a dramatic
impact on the performance on \rv. This is because \wq and \fbq
questions follow simple patterns and are well formed, while \rv
questions have more syntactic and lexical variability. Thus,
paraphrases are important to avoid overfitting on specific question
patterns of the training sets.

\section{Conclusion}

  This paper presents an implementation of MemNNs for the task
  of large-scale simple QA.
  Our results demonstrate that, if properly trained, MemNNs are able
  to handle natural language and a very large memory (millions of
  entries), and hence can reach state-of-the-art on the popular
  benchmark \wq.
%

 We want to emphasize that many of our findings, especially those
 regarding how to format the KB, do not only concern MemNNs but
 potentially any QA system.
 This paper also introduced the new dataset \fbq, which, with $100$k
 examples, is one order of magnitude bigger than \wq: we hope that it
 will foster interesting new research in QA, simple or not.

\bibliographystyle{acl}
\bibliography{qawemb}

\end{document}